\documentclass[runningheads,a4paper]{llncs}

\usepackage{amssymb}
\usepackage{graphicx}

\usepackage{url}

\usepackage{wrapfig}
\usepackage{amsmath}
\usepackage{booktabs}
\usepackage{subcaption} 
\usepackage{color}
\usepackage{hyperref}
\usepackage{multirow}

\usepackage{dsfont} 
\newcommand{\comment}[1]{{\color{red} #1}} 

\newcommand{\Reals}{\mathds{R}}

\usepackage{diagbox} 

\begin{document}

\mainmatter  

\title{Confidence-based Out-of-Distribution Detection: A Comparative Study and Analysis}
\titlerunning{Confidence-based Out-of-Distribution Detection}

\author{Christoph Berger\inst{1}\thanks{c.berger@tum.de} \and
Magdalini Paschali\inst{1}\and \\ Ben Glocker\inst{2} \and
Konstantinos Kamnitsas\inst{2}}
\authorrunning{C. Berger et al.}
%
\institute{ Computer Aided Medical Procedures, Technical University Munich, Germany \and
Department of Computing, Imperial College London, UK}

\toctitle{Lecture Notes in Computer Science}
\tocauthor{Authors' Instructions}
\maketitle



\begin{abstract}
Image classification models deployed in the real world may receive inputs outside the intended data distribution. For critical applications such as clinical decision making, it is important that a model can detect such out-of-distribution (OOD) inputs and express its uncertainty. In this work, we assess the capability of various state-of-the-art approaches for confidence-based OOD detection through a comparative study and in-depth analysis. First, we leverage a computer vision benchmark to reproduce and compare multiple OOD detection methods. We then evaluate their capabilities on the challenging task of disease classification using chest X-rays. Our study shows that high performance in a computer vision task does not directly translate to accuracy in a medical imaging task. We analyse factors that affect performance of the methods between the two tasks. Our results provide useful insights for developing the next generation of OOD detection methods.
\end{abstract}


\section{Introduction}
\label{sec:intro}

Supervised image classification has produced highly accurate models, which can be utilized for challenging fields such as medical imaging. For the deployment of such models in critical applications, their raw classification accuracy does not suffice for their thorough evaluation. 
Specifically, a major flaw of modern classification models is their overconfidence, even for inputs beyond their capacity.
For instance, a model trained to diagnose pneumonia in chest X-rays may have only been trained and tested on X-rays of healthy controls and patients with pneumonia. However, in practice the model may be presented with virtually infinite variations of patient pathologies. In such cases, overly confident models may give a false sense of their competence. Ideally, a classifier should know its capabilities and signal to the user if an input lies out of distribution.

In this work, we first explore confidence- and distance-based approaches for out-of-distribution (OOD) detection on a standard computer vision (CV) task and afterwards evaluate the best OOD detection methods on a medical benchmark dataset. Moreover, we provide a set of useful insights for leveraging OOD approaches from computer vision to challenging medical datasets. 

\textbf{Related work:} 
OOD detection methods can be divided in two categories. The first consists of methods that build a \textbf{dedicated model for OOD detection}~\cite{ruff2020unifying}. 
Some works accomplish this via estimating density $p(x)$ of `normal' in-distribution (ID) data and then classify samples with low $p(x)$ as OOD~\cite{kobyzev2020normalizing} 
However, learning $p(x)$ accurately can be challenging. An alternative is to learn a decision boundary between ID and OOD samples. Methods~\cite{scholkopf2001estimating} attempt this in an unsupervised fashion using only `normal' data. 
Nonetheless, supervised alternatives have also been introduced for CV and medical imaging~\cite{hendrycks2018deep,roy2021does,tan2020detecting}, exposing the OOD classifier to OOD data during training. Such OOD data can originate from another database or be synthesized.   
However, collecting or synthesising samples that capture the heterogeneity of OOD data is challenging. Another approach for creating OOD detection neural networks (NNs) is \emph{reconstruction-based} models~\cite{japkowicz1995novelty,pang2020deep}. 
A model, such as an auto-encoder, is trained with a reconstruction loss using ID data. Then, it is assumed that the reconstruction of unseen OOD samples will fail, thus enabling their detection. This approach is especially popular in medical imaging research~\cite{pawlowski2018unsupervised,you2019unsupervised,baur2021autoencoders,schlegl2017unsupervised,pinaya2021unsupervised},
likely because it produces a per-pixel OOD score, allowing its use for unsupervised segmentation. It has shown promise for localisation of salient abnormalities
but does not reach the performance of supervised models in more challenging tasks.

The second category of OOD detection methods, which this study focuses on, enhances a task-specific model to detect when an input is OOD. These approaches are commonly based on \textbf{confidence of model predictions}.
They are compact, integrated straight into an existing model, and operate in the task-specific feature or output space.
Their biggest theoretical advantage in comparison to training a dedicated OOD detector is that if the main model is unaffected by a change in the data, the OOD detector also remains unaffected. 
A subset of confidence-based methods has a probabilistic motivation, exploring the use of the predictive uncertainty of a model, such as Maximum Class Probability (MCP)~\cite{hendrycks17baseline}, MCDropout~\cite{balcan2016dropout} or ensembling~\cite{lakshminarayanan2017deepensemble}. Others derive confidence-scores based on distance in feature space~\cite{amersfoort2020duq}, or learn spaces that better separate samples via confidence maximization~\cite{liang2018enhancing} or contrastive losses~\cite{amersfoort2020duq,tack2020csi}.
In medical imaging, related work is mostly focused on improving uncertainty estimates by DNNs~\cite{tanno2017bayesian,monteiro2020stochastic}, or analysing quality of uncertainty estimates in \emph{ID} settings~\cite{nair2020exploring,jungo2020analyzing}. In contrast, investigation of OOD detection based on model confidence is limited. A recent study compared MCDropout and ensembling~\cite{mehrtash2020confidence} for medical imaging, finding the latter more beneficial. The potential of other OOD detection methods for medical imaging is yet to be assessed adequately,
despite their importance for the field.

\begin{figure}[h]
\centering
   \includegraphics[width=0.9\linewidth]{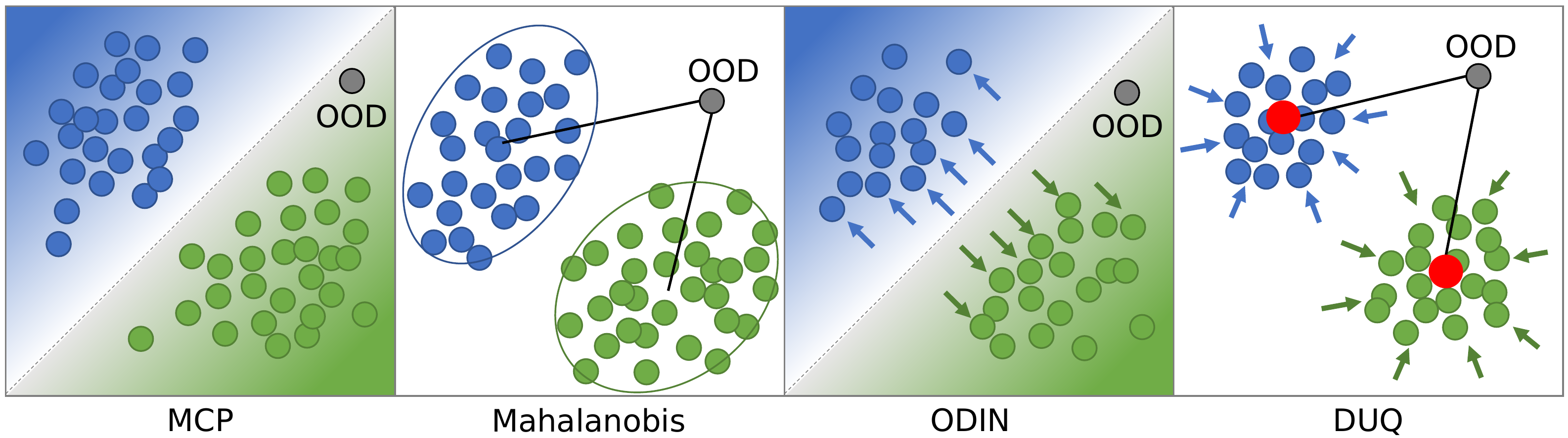}
   \caption{Overview of the OOD detection methods studied: Maximum Class Probability (baseline), Mahalanobis Distance, ODIN and DUQ.}
   \label{fig:Intro} 
\end{figure}

\textbf{Contributions:} This study assesses confidence-based methods for OOD detection. To this end, we re-implement and compare approaches, shown in Figure~\ref{fig:Intro}, in a common test-bed to accomplish a fair and cohesive comparison.
We first evaluate those approaches on a CV benchmark to gain insights for their performance. Then, we benchmark all approaches on real-world chest X-rays~\cite{irvin2019chexpert}. 
We find that the performance of certain methods varies drastically between OOD detection tasks, which raises concerns about their reliability for real-world use, and we identify a method that is consistently high performing across tasks.
 Finally, we conduct an empirical analysis to identify the factors that influence the performance of these methods, providing useful insights towards building the next generation of OOD detection methods.

\section{Out-of-Distribution Detection Methods}
\label{sec:main_methods}

We study the following methods for OOD detection in image classification: 

\noindent\textbf{Maximum Class Probability (MCP) \cite{hendrycks17baseline}:} Any softmax-based model produces an estimate of confidence in its predictions via its class posteriors. Specifically, the probability $\max_y\:p(y|x)$ of the most likely class is interpreted as an ID score and, conversely, low probability indicates possible OOD input. Even though modern NNs have been shown to often produce over-confident softmax outputs~\cite{guo2017calibration}, this method is a useful baseline for OOD detection.

\noindent\textbf{Mahalanobis Distance~\cite{lee2018mahalanobis}:} Lee et al. propose the Mahalanobis distance as OOD metric in combination with NNs. The method can be integrated to any pre-trained classifier. It assumes that the class-conditional distributions of activations $z(x;\theta) \in \Reals^Z$ in the last hidden layer of the pre-trained model follow multivariate Gaussian distributions. After training model parameters $\theta$, the model is applied to all training data to compute for each class $c$, the mean $\hat{\mu}_c \in \Reals^Z$ of activations $z$ over all training samples $x$ of class $c$, and the covariance matrix $\hat{\Sigma}$ of the class-conditional distributions of $z$. To perform OOD detection, the method computes the Mahalanobis distance between a test sample $x$ and the closest class-conditional distribution as follows: 
\begin{equation}
M(x) = {max}_c -(z(x;\theta) - \hat{{\mu}}_c)^{T}\hat{\Sigma}^{-1}(z(x;\theta) - \hat{{\mu}}_c)
\end{equation}
The threshold to decide whether an input is OOD or ID is then set as a certain distance from the closest distribution.

\noindent\textbf{Out-of-Distribution Detector for Neural Networks (ODIN)~\cite{liang2018enhancing}:}
This method is also applicable to pre-trained classifiers which output class-posteriors using a softmax. Assume $f(x;\theta) \in \Reals^C$ are the logits for $C$ classes. We write $S(x,\tau)=\text{softmax}(f(x;\theta), \tau) \in \Reals^C$ for the softmax output calculated for temperature $\tau$ ($\tau_{tr}\!=\!1$ for training), and $S(x,\tau)_c$ is the value for class $c$. The method is based on the assumption that we can find perturbations of the input that increase the model's confidence more drastically for ID samples than for OOD samples.
The perturbed version of input $x$ is given by:
\begin{equation} \label{eq:odin_pert}
\tilde{x} = x -  \varepsilon  sign(- \nabla x \text{ log } max_c S(x;\theta, \tau_{tr})_c) 
\end{equation}

Here, a gradient is computed that maximizes the softmax probability of the most likely class. The model is then applied on the perturbed sample $\tilde{x}$ and outputs softmax probabilities $S(\tilde{x};\tau^{\prime}) \in \Reals^C$. From this, the MCP ID score is derived as $\max_c\:S(x;\tau^{\prime})_c$. Since the perturbation forces over-confident predictions, it negatively affects calibration. To counteract this, ODIN proposes using a different softmax temperature $\tau^{\prime}$ when predicting the perturbed samples, to re-calibrate its predictions. $\tau^{\prime}$ is a hyperparameter that requires tuning. We assess the effect of the perturbation and $\tau$ in an ablation study. 

\noindent\textbf{Deep Ensembles~\cite{lakshminarayanan2017deepensemble}:} This method trains multiple models from scratch, while initialisation 
and order of training data is varied. During inference, predicted posteriors of all models are averaged to compute the ensemble's posteriors. This in turn is used to compute MCP of the ensemble as an ID score. While deep ensembles have been shown to perform well for OOD detection, they come with high computational cost as training and inference times scale linearly with number of ensemble members. In our experiments, we also investigate an ensemble that uses a consensus Mahalanobis distance as OOD score instead of MCP.

\noindent\textbf{Monte Carlo Dropout (MCDP)~\cite{balcan2016dropout}:} MCDP trains a model with dropout. At test time, multiple predictions are made per input with varying dropout masks. The predictions are averaged and MCP is used as ID score. The method interprets these predictions as samples from the model's posterior, where their average is a better predictive uncertainty estimate, improving OOD detection.

\noindent\textbf{Deterministic Uncertainty Quantification (DUQ)~\cite{amersfoort2020duq}:} 
This method trains a feature extractor without a softmax layer. Instead, it learns a centroid per class and attracts samples towards the centroids of their class, similar to contrastive losses \cite{hadsell2006dimensionality}. It uses a Radial Basis Function (RBF) kernel to compute the distance between the input's embedding and the class centroids. The distance to the closest centroid defines classification, and is also used as the OOD score. Because RBF networks are prone to feature collapse, DUQ introduces a gradient penalty to regularize learnt embedding and alleviate the issue. Nonetheless, we still faced difficulties with DUQ convergence despite considerable attempts.

\section{Benchmarking on CIFAR10 vs SVHN}
\label{sec:cifar_svhn}

We first show results on a common computer-vision (CV) benchmark to gain insights about methods' performance, and validate our implementations by replicating results of original works before applying them to a biomedical benchmark.

\subsection{Experimental Setup}
\label{subsec:cifar_setup}

\noindent\textbf{Dataset:} We use the training and test splits of CIFAR10~\cite{cifar10} as ID and SVHN~\cite{svhn2011netzer} as OOD test set ($n_{\text{test\:ID}}\!=\!10000$, $n_{\text{test\:OOD}}\!=\!26032$). 
A random subset of 10\% CIFAR training data is used as validation set, to tune method hyperparameters, such as temperature $\tau$ for ODIN.

\noindent\textbf{Model:} We use a WideResNet (WRN)~\cite{Zagoruyko2016WRN} with depth 28 and widen factor 10 (WRN 28x10), trained with SGD using momentum $0.9$, weight decay $0.0005$, batch normalization and dropout of $0.3$ for 200 epochs with early stopping.

\noindent\textbf{Evaluation Metrics:} We use the following metrics to assess the performance of a method in separating ID from OOD inputs: (1) area under the receiver operating characteristic (AUROC), (2) area under the precision-recall curve (AUCPR), (3) accuracy (Acc) on ID test set. We also use (4) Expected Calibration Error (ECE) as a summary statistic for model calibration~\cite{naeini2015ece}.

\subsection{Results}
\label{subsec:cifar_results}

In Table~\ref{tab:cifar10_overview}, we compare OOD detection performance for all studied methods. MCDP marginally improves over the baseline, with higher gains by Deep Ensembles. Interestingly, ODIN achieves comparable AUROC with Deep Ensembles and ODIN's input perturbation is the component responsible for the performance (see ODIN (pert. only)). The results of only applying temperature scaling and no input perturbation are listed under ODIN (temp. only). The highest AUROC over all methods is achieved by Mahalanobis distance both as a single model and an ensemble. Moreover, none of the OOD detection methods compromised the accuracy on the classification task. 
We reproduced the results of original implementation of DUQ
with ResNet50. However, we faced unstable training of DUQ on our WRN and did not obtain satisfactory performance despite our efforts.

\begin{table}[tbp]
	\begin{center}
	\caption{Out-of-distribution detection performance of WideResNet 28x10 trained on CIFAR10 with SVHN as OOD set. We report averages over 3 seeds.}
	\begin{tabular}{@{}lccc@{}}
	    \toprule
		\textbf{Method} & \textbf{AUROC} & \textbf{AUCPR} & \textbf{ID Acc.}\\
        \midrule 
        MCP (baseline)    & 0.939 & 0.919 & 0.952 \\ 
		MCDP              & 0.945 & 0.919 & \textbf{0.956} \\
		Deep Ensemble     & 0.960 & 0.951 & 0.954  \\
		Mahalanobis       & 0.984 & 0.960 & 0.952 \\
		Mahalanobis Ens.  & \textbf{0.987} & \textbf{0.967} & 0.954 \\
		ODIN              & 0.964 & 0.939 & 0.952 \\
		ODIN (pert. only)  & 0.968 & 0.948 & 0.952 \\
		ODIN (temp. only) & 0.951 & 0.920 & 0.952 \\
		\midrule
        DUQ               & 0.833 & - & 0.890 \\
		\bottomrule
	\end{tabular}
	\label{tab:cifar10_overview}
	\end{center}
\end{table}

\section{Benchmarking on the X-ray Lung Pathology Dataset}
\label{sec:chexpert}

\subsection{Experimental Setup}
\label{subsec:chexpert_setup}

\noindent\textbf{Dataset:} To simulate a realistic OOD detection task in a clinical setting, we use subsets of the CheXpert X-ray lung pathology dataset~\cite{irvin2019chexpert} as ID and OOD data, in two different settings. Since CheXpert images are multi-labeled, we only used samples where ID and OOD classes were mutually exclusive. \textbf{Setting 1}: We train a classifier to distinguish \textit{Cardiomegaly} from \textit{Pneumothorax} (ID), and use images with \textit{Fracture} as OOD ($n_{\text{test\:ID}}=4300$, $n_{\text{test\:OOD}}=7200$). \textbf{Setting 2:} We train a classifier to separate \textit{Lung Opacity} and \textit{Pleural Effusion} (ID), and use \textit{Fracture} and \textit{Pneumonia} as OOD classes ($n_{\text{test\:ID}}=6000$, $n_{\text{test\:OOD}}=8100$).

\noindent\textbf{Model:} We use WRN with depth 100 and a widen factor 2 (WRN 100x2). All other parameters remain the same as for the CIFAR10 vs SVHN benchmark.

\noindent\textbf{Evaluation:} We analyse performance based on the same metrics as in Sec.~\ref{sec:cifar_svhn}.

\subsection{Results}
\label{subsec:chexpert_results}

\begin{table}[t]
	\begin{center}
	\caption{Performance of different methods for separation of out-of-distribution (OOD) from in-distribution (ID) samples for CheXpert in two settings.
	\textbf{Setting 1:} Classifier trained to separate \emph{Cardiomegaly} from \emph{Pneumothorax} (ID) is given samples with \emph{Fractures} (OOD).
	\textbf{Setting 2:} Classifier trained to separate \emph{Lung Opacity} from \emph{Pleural Effusion} (ID) is given samples with \emph{Fracture} or \emph{Pneumonia} (OOD). We report average over 3 seeds per experiment. Best in \textbf{bold}.}
	\begin{tabular}{@{}lccc|ccc@{}}
	    \toprule
		    & \multicolumn{3}{c}{\textbf{Setting 1}} & \multicolumn{3}{c}{\textbf{Setting 2}} \\
		    \cmidrule(lr){2-4} \cmidrule(lr){5-7}
            & \multicolumn{2}{c}{\textbf{OOD}} & \textbf{ID} & \multicolumn{2}{c}{\textbf{OOD}}   & \textbf{ID} \\
		    \cmidrule(lr){2-3} \cmidrule(lr){4-4} \cmidrule(lr){5-6} \cmidrule(lr){7-7}
        \textbf{Method} & \textbf{AUROC} & \textbf{AUCPR} & \textbf{Acc} & \textbf{AUROC} & \textbf{AUCPR} & \textbf{Acc}\\
        \midrule 
		MCP (baseline)   & 0.678 & 0.695 & 0.888                   & 0.458 & 0.586 & 0.758 \\
		MCDP             & 0.696 & 0.703 & 0.880                   & 0.519 & 0.637 & 0.756 \\
		Deep Ensemble    & 0.704 & 0.705 & \textbf{0.895}          & 0.445 & 0.582 & \textbf{0.769} \\
		Mahalanobis      & 0.580 & 0.580 & 0.888                   & 0.526 & 0.601 & 0.758 \\
		Mahalanobis Ens. & 0.596 & 0.586 & \textbf{0.895}          & 0.537 & 0.613 & 0.758 \\
		ODIN             & \textbf{0.841} & \textbf{0.819} & 0.888 & 0.862 & 0.856 & 0.758 \\
		ODIN (pert. only) & \textbf{0.841} & \textbf{0.819} & 0.888 & \textbf{0.865} & \textbf{0.856} & 0.757 \\
		ODIN (temp. only) & 0.678 & 0.695 & 0.888                  & 0.444 & 0.575 & 0.757 \\
		\bottomrule
	\end{tabular}
	\label{tab:chexpert_overview}
	\end{center}
\end{table}

Results for the two ID/OOD settings in CheXpert are shown in Table~\ref{tab:chexpert_overview}. The baseline performance indicates that the ID and OOD inputs are harder to separate for Setting 2, and much harder than the CIFAR vs SVHN task.
MCDP improves OOD detection in both Settings. Interestingly, Deep Ensembles, often considered the most reliable method for OOD detection, do not improve Setting 2, although the Mahalanobis Ensemble does. Moreover, ODIN shows best performance in both settings with a considerable margin, even when only using the adversarial-inspired component of the method without softmax tempering (see ODIN (pert. only) in Figure~\ref{tab:chexpert_overview}). Mahalanobis distance, which was the best method on the CIFAR10 vs SVHN task, performs worse than the Baseline on Setting 1 and only yields modest improvements in Setting 2. Reliability of OOD methods is crucial. Thus, the next section further analyses ODIN and Mahalanobis, to gain insights in 
the consistent performance of ODIN and 
the difference between the CV benchmark and CheXpert Setting 1 that may be causing the
inconsistency of Mahalanobis distance.

\subsection{Further Analysis}
\label{subsec:analysis}

\begin{figure}[t]
	\centering
\begin{subfigure}[b]{0.60\textwidth}
	\centering
	\begin{tabular}{@{}lcc@{}}
	    \toprule
		\textbf{Pooling} & \textbf{Layer shape} & \textbf{AUROC} \\
        \midrule 
        None    & [128,56,56] & \textbf{0.6018}  \\
        4x4, stride=2    & [128,56,56] & 0.5634  \\
        2x2, stride=4    & [128,14,14] & 0.5608  \\
        8x8, stride=1    & [128,14,14] & 0.5508  \\
        1x1, stride=4    & [128,14,14] & 0.545  \\
		\bottomrule
	\end{tabular}
	\caption{Results with dimensionality reduction}
	\label{tab:mahalanobis_ablation}
\end{subfigure}
\begin{subfigure}[b]{0.26\textwidth}
	\centering
	\includegraphics[clip=true, trim=0pt 0pt 0pt 0pt,  width=1.0\textwidth]{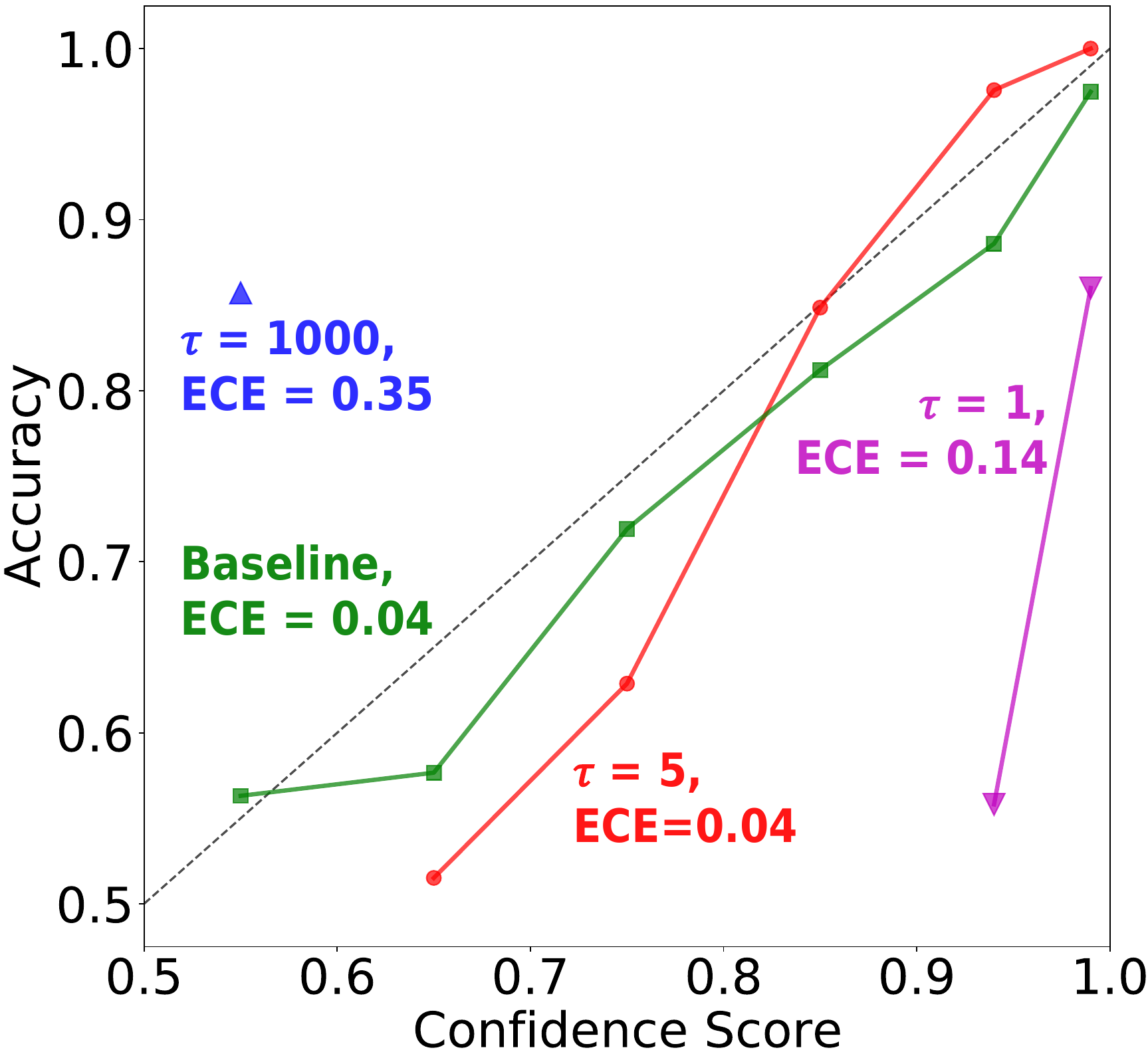}
	\caption{Calibration curves}
	\label{fig:odin_calibration}
\end{subfigure}
\captionsetup{labelformat=empty}
\caption{Table 3a: Results on CheXpert Setting 1 from experiments with dimensionality reduction in last hidden layer. Lower dimensionality did not improve OOD detection via Mahalanobis distance. Fig. 3b: Calibration of baseline and ODIN for varying temperature $\tau$ and associated ECE, for CheXpert Setting 1. The baseline (green) is reasonably calibrated. Adding noise to the inputs with ODIN leads to highly overconfident model (purple, all samples very high confidence). For CheXpert, $\tau\!=\!1000$ as used for CIFAR10 leads to under-confident model, whereas $\tau\!=\!5$ restores good calibration. Interestingly, all ODIN settings achieve the same AUROC for OOD irrespective of $\tau$ value and calibration.
}
\label{fig:ind_vs_transd_TBI_and_examples}
\end{figure}

\noindent\textbf{Mahalanobis:} Our first hypothesis to explain the poor performance of Mahalanobis on the medical OOD detection task in comparison to the CV task was that the Mahalanobis distance may be ineffective in higher dimensional spaces.
In the CIFAR10 vs SVHN task, the Mahalanobis distance is calculated in a hidden layer with [640,8,8] ($40960$ total) activations, whereas the WRN 100x2 for CheXpert has a corresponding layer with shape [128,56,56] ($401408$ total) activations. To test this hypothesis, we reduce the number of dimensions on which we compute the distributions by applying strided max pooling before computing the Mahalanobis distance and report the results in Table~\ref{tab:mahalanobis_ablation}. We find that this dimensionality reduction is not effective and conclude that this is not the major cause of Mahalanobis ineffectiveness in CheXpert.

\begin{figure}[tb]
\centering
\includegraphics[width=1\linewidth]{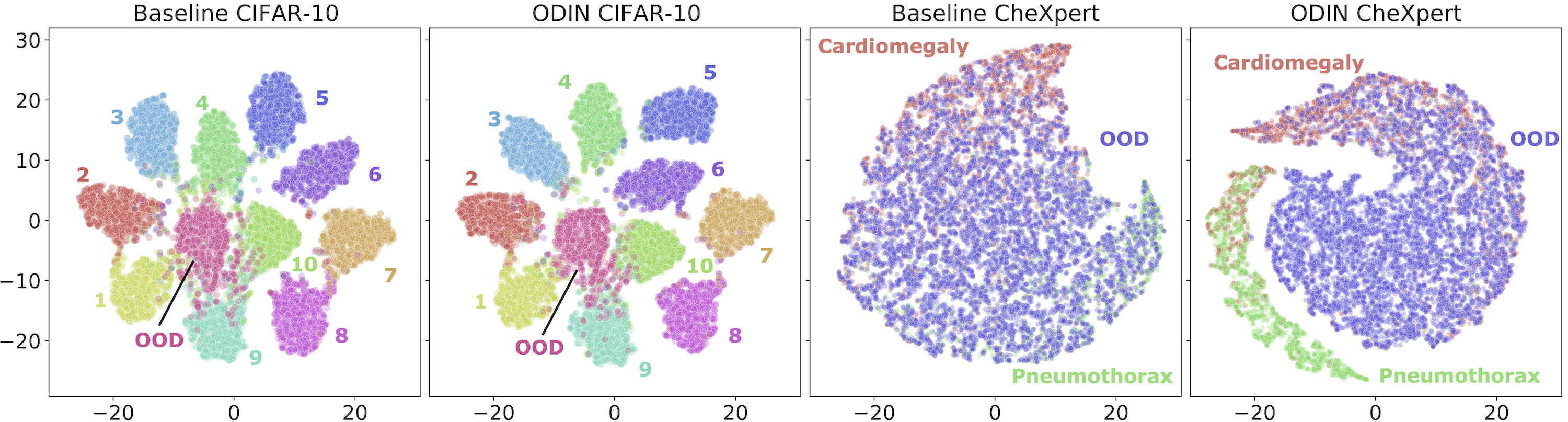}
\caption{
T-SNE of embeddings for CIFAR10 vs SVHN and for CheXpert Setting 1.
The OOD cluster is less separated for the latter, challenging benchmark.
ODIN perturbations improve separation, which may explain its performance.}
\label{fig:tsne}
\end{figure}

To further investigate, we visualize with T-SNE~\cite{vandermaaten2008tsne} the last layer activations when trained models process perturbed samples for the CIFAR10 vs SVHN task and the CheXpert Setting 1. Figure~\ref{fig:tsne} shows that activations for CIFAR10 classes are clearly separated and the OOD set is distinguishable from the ID clusters. For CheXpert, the baseline model achieves less clear separation of the two ID classes and the OOD class overlaps substantially with the ID classes. This suggests that fitting a Gaussian distribution to the ID embeddings is challenging, causing the Mahalanobis distance to not yield significant OOD detection benefits.

\noindent\textbf{ODIN:} We investigate how the perturbation that ODIN adds to inputs benefits OOD detection. For this, we also show T-SNE plots for both CIFAR10 and CheXpert Setting 1 in Figure~\ref{fig:tsne}. The added perturbation results in a better separation of ID classes in both datasets, with the effect more pronounced for CheXpert. While there is still overlap between the \textit{Fracture} OOD class and the \textit{Pneumothorax} ID class, the clusters are more pronounced which ultimately leads to better OOD detection. 
Finally, we investigate the effect of temperature variation in ODIN. Following~\cite{liang2018enhancing}, temperature 1000 was used for CIFAR10 and CheXpert. By comparing baseline, ODIN (temp. only) and (pert. only) on Tables~\ref{tab:cifar10_overview} and~\ref{tab:chexpert_overview}, we find that OOD detection is primarily improved by perturbation, not temperature scaling, especially on CheXpert.
We note, however, that the perturbations lead to a completely over-confident model using training temperature 1, with all predictions having very high confidence (Figure~\ref{fig:odin_calibration}). AUROC and AUCPR are calculated via ordering the OOD score (i.e. confidence) of predictions, so even slight differences between ID and OOD samples suffice to separate false and true detections. If only those metrics were taken into account, temperature scaling might have been considered redundant.
However, to deploy an OOD system, a threshold on the confidence / OOD score needs to be chosen. Spreading the confidence estimates via temperature scaling ($\tau\!=\!5$ in Figure~\ref{fig:odin_calibration})
enables more reliable choice and deployment of a confidence threshold in practical settings.

\section{Conclusion}
\label{sec:conclusion}
This work presented an analysis of various state-of-the-art methods for confidence-based OOD detection on a computer vision and a medical imaging task. Our comprehensive evaluation showed that the performance of methods in a computer vision task does not directly translate to high performance on a medical imaging task, emphasized by the analysis of the Mahalanobis method. Therefore, care must be given when a method is chosen. We also identified ODIN as a consistently beneficial OOD detection method for both tasks. Our analysis showed that its effect can be attributed to its input perturbation, which enhances separation of ID and OOD samples. This insight could lead to further advances that exploit this property. 
Future work should further evaluate OOD detection methods across other datasets and tasks to better understand which factors affect their performance and reliability towards real-world deployment.

\section*{Acknowledgements}
This work received funding from the European Research Council (ERC) under the European Union's Horizon 2020 research and innovation programme (grant agreement No 757173, project MIRA, ERC-2017-STG), and the UKRI London Medical Imaging \& Artificial Intelligence Centre for Value Based Healthcare.

\bibliographystyle{splncs03}
\bibliography{references}

\end{document}